# Gender Inequality in English Textbooks Around the World: an NLP Approach

Tairan Liu

## Abstract

*Textbooks are important for shaping children's understanding of the world. Previous studies of individual countries have suggested that gender inequality exists. There lacks a study that compares gender inequality in textbooks around the world. This study uses NLP approaches to quantify gender inequality in English textbooks in 7 cultural spheres, 22 countries, by measuring the count, firstness, and TF IDF words by gender. The study also counted the names that appeared in TF IDF word lists and sorted the names by gender, found out that LLMs can distinguish between the different TF IDF word lists, and mapped the TF IDF words to GloVe to see that some keywords are closer to one gender than the other. The study found more male count, firstness, and names. The study found that there is significant gender inequality in all the textbooks. Gender inequality is demonstrated the least in textbooks of the Latin Cultural Sphere.*

## Introduction

Textbooks play a central role in shaping children's understanding of society, identity, and values, acting as mirrors of cultural norms and expectations. Thus, if textbooks convey gender inequality, kids will likely absorb such ideas and enhance the gender inequality of human society. Thus, it is important to examine gender inequality in textbooks, as it is the first step to diminish gender inequality in textbooks.

On top of that, culture can have an influence on the gender inequality contents in textbooks. Although gender inequality is a worldwide problem, it might be more profound of a problem in some cultural spheres than others. Textbooks from countries within a cultural sphere that emphasize on gender roles might depict more gender inequality.

There are a lot of studies done on textbooks that teach English of individual countries that have found gender inequality. English textbooks from China (Zhang et al., 2022), Mexico (Sánchez Aguilar, 2021), Saudi Arabia (Sulaimani, 2017), Vietnam (Vu & Pham, 2023), the Philippines (Tarrayo, 2014), and India (Bhattacharya, 2017) have all portrayed gender inequality. In many of these studies, the inequality of social roles is pointed out. When female and male characters appear in the textbook, chances are that they do not work in the same occupation. Males might have more high end jobs, while females get more family related jobs (Bhattacharya, 2017; Sánchez Aguilar, 2021; Tarrayo, 2014; Vu & Pham, 2023; Zhang et al., 2022). Also, studies have shown that there is an overall more male presence (Bhattacharya, 2017; Sánchez

Aguilar, 2021; Sulaimani, 2017; Tarrayo, 2014; Vu & Pham, 2023) and firstness (Bhattacharya, 2017; Tarrayo, 2014), which is that when characters of both genders appear in tandem, males are usually mentioned first. Several studies have also indicated that males are described in more positive terms (Bhattacharya, 2017; Tarrayo, 2014), and both genders are described in words that fit into the stereotypical gender image (Bhattacharya, 2017; Tarrayo, 2014; Vu & Pham, 2023).

In order to identify gender inequality in textbooks, some traditional methods include manual qualitative analysis. However, while there will be both qualitative and quantitative results generated from this study, this study uses automated large corpus analysis methods instead of manual analysis. This study uses Natural Language Processing (NLP) methods to quantify inequality between genders in the textbooks.

Furthermore, while previous studies on gender inequality of textbooks have been quite abundant, they are mostly about analysing the gender inequality of textbooks from one specific country or comparison between textbooks that are from different countries in the same cultural sphere. This study will take a step further by comparing the gender inequality between textbooks from different cultural spheres.

Since NLP analysis methods are used, this research will only go into English textbooks to control the confounding variable of language. Here are the hypothesis that will be tested in this study:
1. Gender inequality exists in English textbooks in general.
2. Gender inequality is presented differently in the English textbooks, depending on the cultural sphere that the source country of the textbooks.

The way that this study divides up cultural spheres is adopted is from Huntington's *The Clash of Civilizations*. The cultural spheres include:
1. Sinosphere: influenced by Chinese culture
2. Indosphere: influenced by Indian Culture
3. Islamic cultural sphere: muslim countries that are in middle east and North Africa
4. African cultural sphere: Mainly in the South of Africa
5. Western cultural sphere: West Europe, North America, and some Oceania countries
6. Eastern Europe cultural sphere: East Europe
7. Latin American cultural sphere: Mexico and Most countries in South America.(Huntington, 1993)

There have been previous studies that demonstrated that LLMs have hidden gender bias due to its capturing of the gender biases in the corpus that it is trained on. Although LLM companies have made the effort to regulate the gender biased speech of their LLM models, the gender bias in LLMs are ingrained and can be elicited (Dong et al., 2023).

GloVe is created by global co-occurrence between different words (Pennington et al., 2014), so it captures the gender bias in the real world (Dawkins, 2023).

# Methods

In order to quantify the hypothesis, the study will explore gender inequality through several different aspects. First, this study looks to count the occurrences and firstness of both genders. Count stands for the count of appearance of each gender, and firstness means that when mentioned together, which gender appears first. For example, "ladies and gentleman" is a female first occasion, whereas "boys and girls" is a male first occasion. On top of that, this study will seek for the keywords associated with both genders using Term Frequency-Inverse Document Frequency. The word lists themselves can serve as a qualitative result. The qualitative result will then be quantified by counting the number of names, LLM recognition, and Distance to Keywords. Counting the number of names means to count the occurrence of names of different genders from the TF IDF word lists. LLM recognition testifies if LLM are able to attribute the TF IDF word lists to their corresponding gender. Distance to Keywords calculates the distance of the TF IDF word lists to several selected keywords on GloVe.

## Data Gathering and Grouping

This study focuses on English textbooks that are used in 7-9th grade, which is middle school for most countries around the world. Books for young children are not appropriate for the study because they are more graphic and less textual, giving less content to analyze on. Also, it will be difficult to find textbooks that are used in high school and college, as a lot of countries do not require English teaching for that age range. Thus, this study will only focus on English textbooks used in 7-9th grade.

One way of selecting textbooks is through asking locals. Textbooks of 10 countries are collected this way. There are a number of countries that publish the list of textbooks they use for all of their middle schools, and there are also some schools that publish the list of textbooks they use since they are not designated by their government. These account for textbooks of the rest of the countries collected for this study. In total, English textbooks from 23 countries and regions are collected. Here are the textbooks with their corresponding country and cultural sphere:

| Country | Title | Publisher | Cultural sphere |
| --- | --- | --- | --- |
| Argentina | *For Teens* | Pearson Longman | Latin American Cultural sphere |
| Australia | *English for Academic Purposes (EAP)* | Pearson Longman | Western Cultural sphere |

| Belarus | *English* | Higher School Publishing House | East Europe Cultural sphere |
|---|---|---|---|
| China | *English* | Education Science Electron Publishing House | Sinosphere |
| Germany | *Green Line* | Klett | Western cultural sphere |
| India | *Honey Comb* | National Council of Educational Research and Training | Indosphere |
| Iran | *Prospect* | Iranian Publication | Islamic cultural sphere |
| Japan | *New Horizon* | Tokyo Shoseki | Sinosphere |
| Malaysia | *English* | Kementerian Pendidikan Malaysia | Indosphere |
| Mexico | *Yes, We Can!* | Richmond | Latin American sphere |
| Philippians | *Learning Package* | Philippians Department of Education | Western Cultural sphere |
| Saudi Arabia | *Super Goal* | McGraw Hill | Islamic cultural sphere |
| Southern Cone | *New Opportunities* | Pearson Longman | Latin American Cultural Sphere |
| | *New Headway* | Oxford | |
| | *Pacesetter* | Oxford | |
| | *Click On* | Express Publishing | |
| | *New Snapshot* | Pearson Longman | |
| South Sudan | *Secondary English* | Ministry of general education and instruction | African Cultural Sphere |
| Spain | *English In Mind* | Cambridge | Western cultural sphere |

| Thailand | *English in Daily Life* | Office of non-formal and Informal Education Promotion, office of the Permanent Secretary of the Ministry of Education, Ministry of Education | Indosphere |
| --- | --- | --- | --- |
| Tunisia | *Let's Discover More English* | National Pedagogic Centre | Islamic cultural sphere |
| Turkey | *Uplift* | State Books | East Europe sphere |
| UK | *Touchstone* | Cambridge | Western cultural sphere |
| Ukraine | *Streamline English Departure* | Oxford English | East Europe Cultural Sphere |
| Uzbekistan | *Fly High* | TOSHKENT «Yangiyo'l poligraf servis» | East Europe Cultural Sphere |
| Vietnam | *Tiếńg Anh* | Vietnam Education Publishing House | Sinosphere |

*Table 1: List of textbooks used for the study in alphabetical order.*

## Data Cleaning

Since the data collected previously are in PDF format, the first step of data cleaning is to extract the textual data from the PDF. If the texts from the PDF can be extracted directly, they are read in directly using the fitz method from PyMuPDF. However, if the PDF is stored in an image format, meaning that the text cannot be extracted directly. In this case, OCR from Adobe can help convert the PDF to identifiable text, allowing it to be then read in by fitz. The texts are cleaned using NLTK to lowercase, remove punctuation, and remove stopwords. LangDetect is used to remove everything that is not English.

Then, according to the country of origin, texts of the same cultural sphere are put into one list. The 7 cultural spheres are set aside for the first round of analysis, since the first round of analysis is done on all the textbooks.

The text is then divided into segments of 100 characters, and if a gendered keyword occurs in the segment, the segment of text is saved into the list of the corresponding document of that gender. If it has keywords from both genders, it is saved into both documents. This example (which has stopwords and punctuation removed) will be put in the female document because it contains a female keyword "mother": "important politician country complain injustice assuming mother happen important politician course". Here is the list of keywords:

| Male keywords | Female Keywords |
|---|---|
| men | women |
| boy | girl |
| male | female |
| brother | sister |
| father | mother |
| son | daughter |
| husband | wife |
| king | queen |
| prince | princess |
| uncle | aunt |
| nephew | niece |
| he | she |
| him | her |
| his | hers |
| gentleman | lady |
| sir | Ma'am or madam |
| mr. | Mrs. or ms. or miss |
| hero | heroine |
| lord | dame |
| patriarch | matriarch |
| man | woman |

*Table 2: List of gender keywords used for locating gendered contexts.*

The same is done to texts of each individual cultural sphere, so that there are two documents for each cultural sphere, and there are two documents for all the textbooks as a whole.

## Count

This study counts the number of segments in both documents. This measures the number of occurrences of segments that contain gendered keywords. In texts without gender bias, the counts should be either the same or have insignificant differences. Thus, it would be fair to assume that it should be a Bernoulli distribution with a probability of 0.5. A p value lower than 0.01 under this situation would signify gender inequality.

## Firstness

Firstness is a frequently used method to assess gender inequality. Firstness refers to that, when two genders are mentioned at the same time, which gender comes first. For example, "ladies and gentleman" would be a female first occurrence, and "boys and girls" would be a male first occurrence.

It is noteworthy that firstness is only counted when the two gendered words are of the same level. As shown in Table 2, only words of the same row would be considered the same level. For example, while "father and mother" is a male first occurrence, "mother and son" is not a female first occurrence. This is because "father" and "mother" are on the same level, same row of Table 2, but "mother" and "son" are not of the same level. In previous studies on gender inequality in textbooks, this factor is put into consideration when calculating firstness (Bhattacharya, 2017; Tarrayo, 2014).

In texts without gender bias, the counts should be either the same or have insignificant differences. Thus, it would be fair to assume that it should be a Bernoulli distribution with a probability of 0.5. A p value lower than 0.01 under this situation would signify gender inequality.

## TF IDF

Term Frequency - Inverse Document Frequency (TF IDF) is conducted to the two documents.

$$TF(t, d) = \frac{number\ of\ times\ t\ appears\ in\ d}{total\ number\ of\ terms\ in\ d}$$

$$IDF(t) = log \frac{N}{1 + df}$$

$$TF - IDF(t, d) = TF(t, d) * IDF(t)$$

In this equation, t refers to the term or the word being evaluated, d refers to a specific document within the corpus, N is the total number of documents in the corpus, and df is the document frequency, which is the number of documents that contain the term t. TF stands for the number of times that a term appears in a document over the total number of terms in the document. IDF stands for log of the total number of documents in the corpus over the number of documents containing the term. Thus, TF IDF reflects both how frequently a term appears in a document and how important it is across all the documents. The higher the TF IDF score, the more important the term is to that document relative to the entire corpus.

In this study, TF IDF is adopted instead of simply frequency because the words that tend to have a high frequency in general will likely have high frequency in both documents. Although stopwords are filtered out, words like "something" might still have high frequency in both documents. Thus, TF IDF is used so that words that have high frequency in both documents can be crossed out, leaving behind the words that are representative of the document compared to the other.

Since TF IDF tends to treat all the words differently, prior to TF IDF, all the words are stemmed using porter. Top 300 words from both lists are taken. Also, because the documents are compiled by segments that contain a keyword, it is reasonable to take out the gendered keywords from the TF IDF word lists. Finally, words that appeared in both TF IDF lists are removed from both lists to get the most unique words.

### Counting Names

After obtaining the two TF IDF lists, the count of names are counted manually. In both lists, names are counted and categorized according to their gender. This gives four categories of names: Male names mentioned in the female TF IDF list, female names mentioned in the female TF IDF list, male names mentioned in the male TF IDF list, and female names mentioned in the male TF IDF list.

### LLM recognition

There are four LLMs that are used to try to distinguish between these two lists: ChatGPT 4.0, Deepseek R1, Gemini 2.0 Flash, and Claude 3.7 Sonnet. The prompt used is as follows: "I did TF IDF to a corpus that consists of 2 documents. One document is in a female context, and the other one is on male context. Here are my two TF IDF list results. Can you guess which one is for female context, which one is for male context? [male context list] [female context list]." Each LLM is tested 3 times in different tabs. For the lists that are used, names are removed.

### Distance to Keywords on GloVe

As a tool that captures global co-occurrence between different words (Pennington et al., 2014), GloVe vectors can be used as an indicator of gender bias in the real world (Dawkins, 2023). Compared to LLMs, GloVe is an older tool that is trained unsupervised, and has words represented as vectors in a vector space. Thus, vector calculations can be done to words, like "king" - "man" + "woman" = "queen" in GloVe vector space (Ethayarajh et al., 2019). If certain

keywords are closer to one TF IDF word list compared to another in the GloVe vector space, it would mean that there is a co-occurrence between the gender and the keyword inferred by the textbooks.

The two lists of TF IDF words are plotted out in the GloVe vector space, and the cosine distance between the lists to some keywords are measured. Here is a list of keywords tested.
- Death
- Food
- Baby
- Pretty
- Love
- Violence
- Household
- Doctor
- Nurse

# Results

In each part of the results portion, both the results obtained from all the textbooks as a whole and the results obtained from each cultural sphere will be reported. They will be reported separately.

The cultural spheres will be written in abbreviated form in the tables. Table 3 is what each abbreviation refers to.

| Abbreviation | Full Meaning |
| --- | --- |
| African | African Cultural Sphere |
| East European | Eastern European Sphere |
| Indosphere | Indosphere |
| Islamic | Islamic Cultural Sphere |
| Latin | Latin Cultural Sphere |
| Sinosphere | Sinosphere |
| West European | Western European Sphere |

*Table 3: List of abbreviations of Cultural Spheres used for this study*

## Count

| Cultural sphere | Female count | Male count | P-value |
|---|---|---|---|
| Overall | 3740 | 4764 | 1.19 * 10^-28 |
| African | 608 | 728 | 0.00224 |
| East European | 342 | 429 | 0.00258 |
| Indosphere | 245 | 430 | 1.03 * 10^-6 |
| Islamic | 246 | 305 | 0.00992 |
| Latin | 914 | 953 | 0.2702 |
| Sinosphere | 307 | 352 | 0.0341 |
| West European | 1078 | 1567 | 1.17 * 10^-20 |

*Table 4: Count of occurrences of the two genders in the textbooks. To see this in a bar chart, refer to Image 1 below.*

Table 4 includes the counts of both all the textbooks as a whole and each cultural sphere and their corresponding p values. The p values are two tails, calculated by taking from z scores of bernoulli with a probability of 0.5, since the count of males and females should be even if gender bias does not exist.

As we can see, most of the cultural spheres and all the textbooks as a whole have a statistically significant difference between the counts. The only two exceptions are Sinosphere and Latin Cultural Sphere. Although there is still a higher male presence in these two cultural spheres, the difference is not statistically significant, with p values of 0.086 and 0.368 respectively.

In Image 1, the red line represents equal occurrence between the two genders. As it clearly demonstrated, there is more of a male occurrence in all of the cultural spheres, as well as overall. The division between the genders in the Latin Culture Sphere is the closest to the red line, meaning that it is more of an equal occurrence compared to the other cultural spheres.

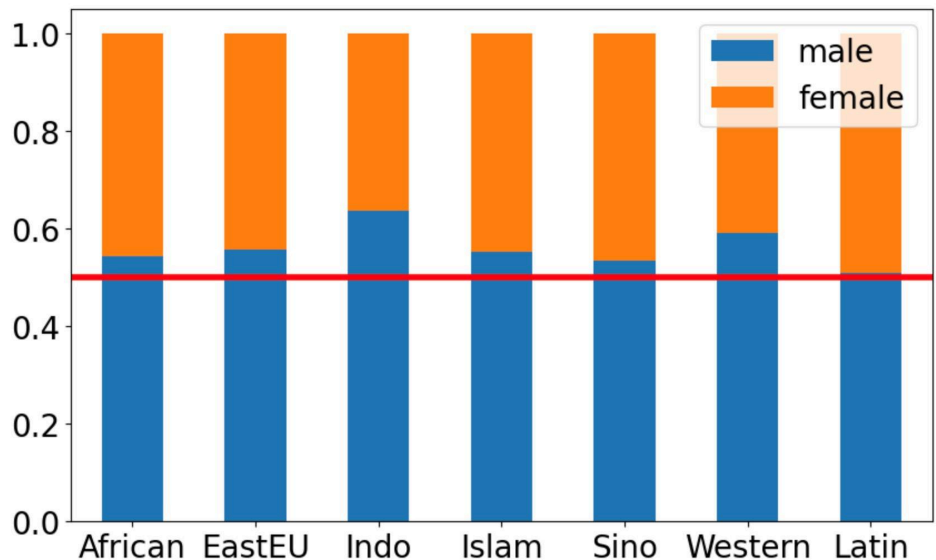

*Image 1: Count of occurrences of the two genders in the textbooks. To see this in a table, refer to Table 4 above.*

## Firstness

| Cultural sphere | Female First | Male First | P-value |
|---|---|---|---|
| Overall | 183 | 247 | 0.00234 |
| African | 38 | 46 | 0.45 |
| East European | 3 | 18 | 0.00149 |
| Indosphere | 4 | 11 | 0.12 |
| Islamic | 10 | 23 | 0.03508 |
| Latin | 87 | 94 | 0.66 |
| Sinosphere | 3 | 9 | 0.15 |
| West European | 38 | 46 | 0.45 |

*Table 5: Count of firstness per gender when two genders are mentioned together. To see this in a bar graph, see Image 2 below.*

Table 5 is about firstness between the two genders in different textbooks. Just like count, there is always more male first than female first in all the cultural spheres and overall. However, only several ones are statistically significant. The overall male firstness is significantly more than the overall female firstness (p<0.01). Firstness is also significant for both the East European

Cultural Sphere (p<0.01) and Islamic Cultural Sphere (p<0.05). However, for all the other cultural spheres, the difference is not significant (p>0.05). This might be due to the lack of firstness occurrence. For example, there are only 12 recorded firstness occurrences in the Sinosphere, and 15 in indosphere.

The culture sphere that has the smallest gender firstness imbalance is the Latin cultural sphere, and the cultural sphere that has the biggest gender firstness imbalance is the East European Cultural sphere.

In Image 2, the red line represents equal occurrence of firstness between the two genders. As it clearly demonstrated, there is more of a male firstness occurrence in all of the cultural spheres, as well as overall. The division between the genders in the Latin Culture Sphere is the closest to the red line, meaning that it is more of an equal occurrence compared to the other cultural spheres.

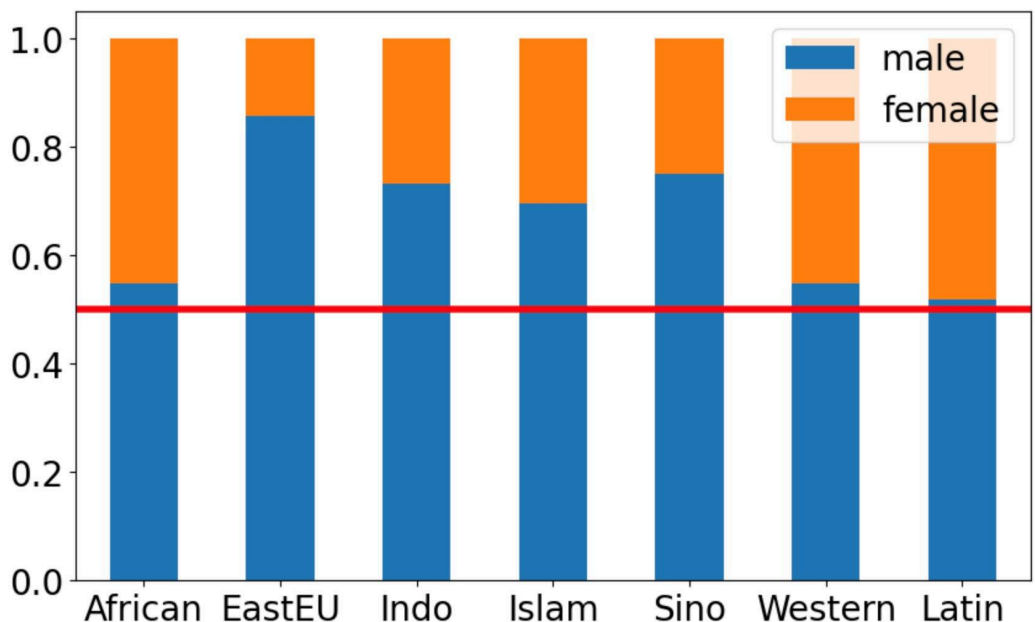

*Image 2: Count of firstness per gender when two genders are mentioned together. To see this in a table, see Table 5 above.*

## TF IDF

### Counting Names

In this section, the names obtained from TF IDF lists will be discussed in the order of female context and male context. The names obtained from the female context are the manually found names in the TF IDF list from the compiled female document as the document, both the female and male document as the corpus. The names obtained from the male context are the manually found names in the TF IDF list from the compiled male document as the document, both the female and male document as the corpus.

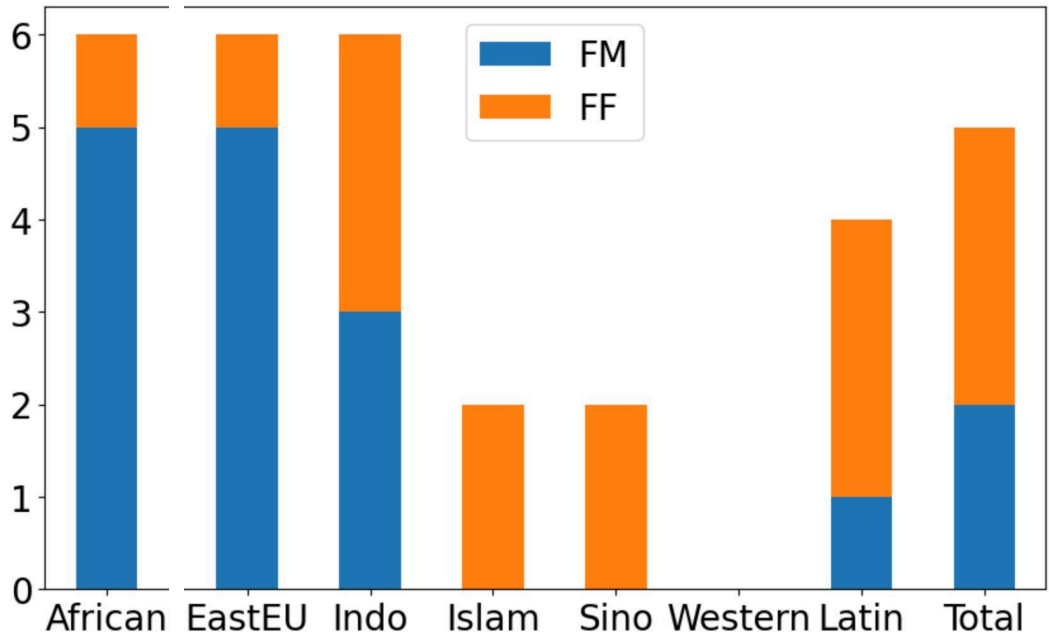

*Image 3: Names in TF IDF list of Female context (FM: Female context, male names, FF: Female context, female names)*

      Image 3 demonstrates the number of names mentioned in the female context TF IDF list. As we can see, in African and East European cultural spheres, there are more male names (5) than female names (1) even in the female context. There are an equal number of male and female names in the Indosphere, and no names in the Western Cultural sphere. For the other cultural spheres, there are more female names than male names.

      Notice how the number of names from different cultures don't add up to the total column. This is because the Total column doesn't represent the number of names from different cultural spheres added up, but it is the number of names from the TF IDF done on the female context of all the textbooks. In this scenario, there are more female names (3) than male names (2).

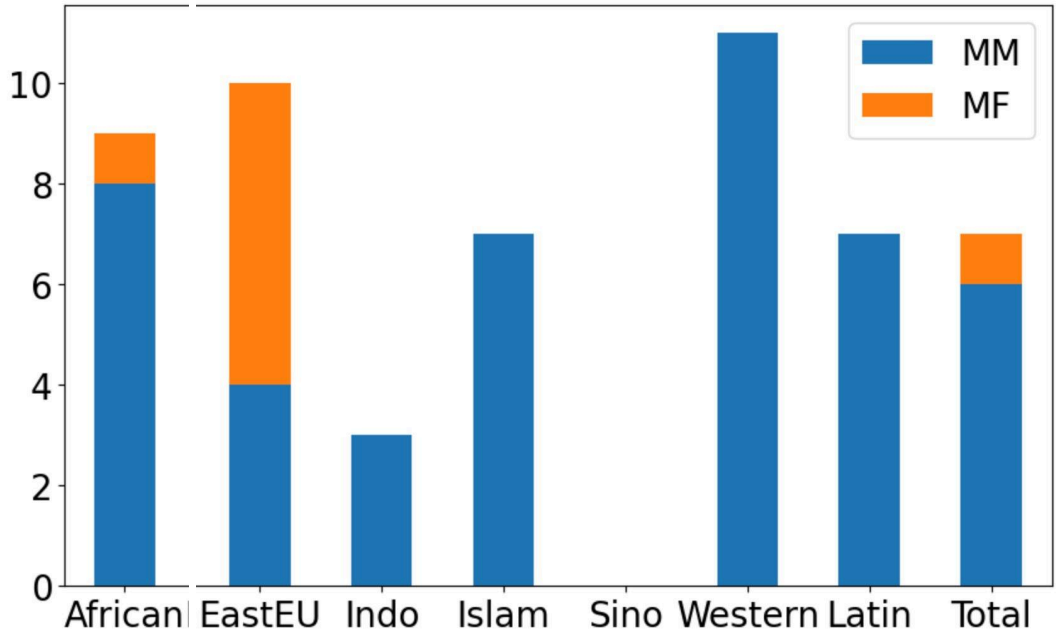

*Image 4: Names in TF IDF list of Male context (MM: Male context, male names, MF: Male context, female names)*

      Image 4 demonstrates the number of names mentioned in the male context TF IDF list. Notice that although the X axis, the name of the cultural spheres, stays the same, the Y axis, the number of names, changed. There are overall more names mentioned in the male TF IDF lists.

      The East European Cultural sphere is the only one that has more female names (6) than male names (4). African and East European Cultural spheres are the only cultural spheres with female names. There are no names in Sinosphere. In Western Cultural sphere, there are 11 male names.When all the textbooks are considered as a whole, there are more male names (5) than female names (1).

      Comparing them across male and female context, we can see that patterns vary between cultural spheres. For the African Cultural sphere, there are always more male names than female names, regardless of the gender context. For the Indosphere, Latin Cultural Sphere, and Islamic Cultural Sphere, there are more female names in the female context, and more male names in male context. For the East European Cultural sphere, it is the opposite: more male names in female context, more female names in male context. For Sinosphere, while there are some female names (2) and no male names in the female context, there are no names in the male context. On the other hand, Western Cultural sphere has some male names (11) and no female names in the male cultural sphere, there are no names in the female context. As a whole, there are more female names (3) than male names in the female context (2), and more male names (5) than female names (1) in the male context.

## LLM Recognition

|  | ChatGPT 4.0 | Deepseek R1 | Claude 3.7 Sonnet | Gemini 2.0Flash |
|---|---|---|---|---|
| Overall | 100% | 100% | 100% | 66% |
| African | 100% | 100% | 100% | 100% |
| East European | 100% | 100% | 100% | 100% |
| Indosphere | 100% | 100% | 100% | 100% |
| Islamic | 100% | 100% | 100% | 66% |
| Latin | 100% | 66%* | 100% | 66% |
| Sinosphere | 100% | 100% | 100% | 100% |
| West European | 100% | 100% | 100% | 100% |

*Table 6: LLM Recognition Accuracy*
*\*: Deepseek is able to tell them apart correctly, but it refers to the first list as the second list and vice versa.*

Overall, the LLMs are able to distinguish between the two lists correctly, despite Deepseek and Gemini having several mistakes on some of the trials (the only three 66% in Table 6). It is important to note that on the trial that Deepseek got wrong, it is not completely wrong. It just keeps referring to words from the first list, talking about how they relate to masculinity, and then proceeds to conclude that the second list is the male list. The deduction procedure is not incorrect, it is just the conclusions are off. Gemini got one trial wrong for all the words, Islamic cultural sphere, and Latin cultural sphere.

## Distance to Keywords

| Keywords closer to Female Cluster | Keywords closer to Male Cluster |
|---|---|
| Pretty | Death |
| Baby | Violence |
| Love |  |
| Nurse |  |

| Food |  |
|---|---|
| Doctor |  |

Table 7: Keywords and the cluster they are closer to, ranked by the difference in distance

|  | Distance to Female Cluster | Distance to Male Cluster | Difference in distance |
|---|---|---|---|
| Death | 0.83 | 0.79 | 0.04 |
| Food | 0.84 | 0.85 | -0.01 |
| Baby | 0.79 | 0.83 | -0.04 |
| Pretty | 0.76 | 0.81 | -0.05 |
| Love | 0.75 | 0.78 | -0.03 |
| Violence | 0.87 | 0.86 | 0.01 |
| Household | 0.89 | 0.90 | -0.01 |
| Doctor | 0.82 | 0.83 | -0.01 |
| Nurse | 0.87 | 0.90 | -0.03 |

Table 8: Distance of Keywords to Gendered Clusters

Although there are several words for both genders, it is important to take note that in table 7, only "pretty", "baby", "love", and "nurse" have more or equal to 0.03 difference for females, and only "Death" for males. Although the rest of the words are closer to the female cluster, they only have 0.01 difference between the distances.

Note that in table 8, the Difference in distance is calculated by the cosine distance to Female Cluster minus the cosine distance to Male Cluster. That means if the difference is positive, it is closer to the male cluster, and thus more associated with males. On the other hand, if the difference is negative, it is closer to the female cluster, and thus more associated with females.

# Discussion

In the first indicator, count, most cultural spheres demonstrated significant differences in the number of occurrences between males and females. The difference is not significant in only the Latin Cultural Sphere and the Sinosphere. This could mean that in these two cultural spheres, the presence of both genders are acknowledged to the same level, although it cannot tell anything about the context when different genders are mentioned. In all the other cultural spheres, the presence of two genders are not acknowledged, with a significant male presence acknowledgement.

In the second indicator, firstness, most cultural spheres have no significant differences between the number of firstnesses. Although there is more male firstness in all the cultural spheres, the number of firstness is insufficient to prove statistically significant. The only two cultural spheres that can show statistical significance are Islamic cultural sphere and East European Cultural sphere. However, it is important to note that when all the textbooks are taken into account for, there is a statistically significant difference between firstness. Thus, the lack of significance in the difference in firstness might disappear if given enough data from each cultural sphere.

In both firstness and count, the Latin American Cultural Sphere has done exceptionally well. Not only does it have the closest ratio between genders for count, it also has the closest ratio for firstness. This is quite surprising, since users of the language Spanish tend to use the masculine form for everything (Beatty‑Martínez et al., 2021). One way to explain this is that the English textbooks adapted by the countries of Latin Cultural Spheres are mostly not from presses that are based in Spanish speaking countries. Instead, they are all presses based in the countries of the Western Cultural Sphere. It is quite ironic that the textbooks adapted from the western cultural sphere, however, are doing worse on these two indices.

For the names found in TF IDF lists, it is important to note that there are more names in the male context (54) than in the female context (31). On top of that, there are more male names (62) than the female names (23). In the female context, there are more male names (16) than female names (15), and in the male context, there are also more male names (46) than female names (8). In order to see the distribution of names according to context, I did a monte carlos on the corpus consisting of the two gendered documents. It turns out that there are no names. This means that all names are significant, but it cannot state anything about the names in different gendered contexts.

For cultural spheres that have more names of the gender corresponding to the context, it would mean that the characters tend to appear on their own or with others of the same gender. Islamic cultural sphere, Sinosphere, Western cultural sphere, and Latin Cultural sphere have this pattern. It is important to note that the Sinosphere only has 2 female names and no male names in the female context, no names in the male context, and Western cultural sphere has no names in the female context, but 11 male names and no female names in the male context. This could mean that the interaction between two female characters is more frequent than other interactions in the Sinosphere textbooks. On the other hand, Western Cultural sphere textbooks focus on the interaction between 11 male characters.

For cultural spheres that have more names of the different gender corresponding to the gender of the context, it might mean that when the gender of the context shows up, it is usually accompanied by characters of another gender. In the African cultural sphere, there are more male names in both gendered contexts. This could mean that while when females are mentioned, they are associated with males, but when males are mentioned, they are less associated with females. This could also be the same case for Indosphere, as there are the same number of names of both genders mentioned in the female context, but only male names are mentioned in the male context. In the East European Cultural Sphere, there are more male names in the female context, and more female names in the male context. This might signify that characters of both genders intertwine when they show up.

It is quite surprising that LLMs are able to distinguish between female and male TF IDF words so well. The only four trials that error occurred in this task are all on the first trial. Although a different chat is used for each trial and feedback is not given to the LLMs after each trial, the memory of the previous query could have been taken into account by the LLM, and thus perform better afterwards.

When plotting the two lists of TF IDF words onto the GloVe space, "pretty", "baby", "love", and "nurse" have more or equal to 0.03 difference for females, and "death" for males. As a tool that captures the gender bias in the world, GloVe proves to us that some words that are closer to some gender in the world of the large corpus are also closer to the TF IDF words according to their gender. This tells that not only are textbooks biased, they are also biased in the same way as the general corpus of the world.

# Conclusion

In conclusion, gender inequality is pervasive in English textbooks all around the world, in all cultural spheres. The study found more male count, firstness, and names.The study also found out that LLMs can distinguish between the different TF IDF word lists, and mapped the TF IDF words to GloVe to see that some keywords are closer to one gender than the other. There are textbooks with relatively less gender inequality, such as those from the Latin American cultural sphere. However, there is unfortunately an overall trend of gender inequality in English textbooks around the world.

There are several limitations to this study. First, there are only three countries accounted for per cultural sphere. This might have caused some insignificant differences in some of the indices, such as the firstness. If there were more textbooks from more countries, there might be more significant differences in indices like firstness. Thus, future study that wishes to replicate this study can be done using more textbooks from each cultural sphere.

Another limitation this study might have is the selection of textbooks. It is easier to obtain textbooks from countries where the government either posts the textbooks for public access or a list of textbooks that are adopted. Thus, during the data collection phase of the study, it is inevitable that more textbooks from such countries are collected. However, since this holds true for all of the cultural spheres, the effect should in theory cancel itself out.

One last limitation is that, in the firstness indicator, not all firstness are assessed. This study calculates firstness strictly in one situation: when one gendered word appears and another gendered word of the opposite gender either appears immediately after or only has one stopword in between. For example, "ladies and gentleman" will be taken into account, since "ladies" is followed by "gentleman" with only one stopword ("and") in between. On the other hand, "boys are doing this. Girls are doing that." is not recorded as a male first occurrence, since the female keyword ("girls") is not following the male keyword ("boys") immediately or with only one stopword in between. However, since this case does not account for both genders, the effect should in theory cancel itself out.

Future work in this field can either go in the direction of expanding the types of textbooks used for analysis and the methods used for analysis. In this study, English textbooks are discussed to control the language as a potential confounding variable for NLP analysis, but future research on NLP might improve and the language barrier might be removed. In such a sense, other textbooks can be analyzed as well, such as math and history. This will be an important step to diminishing gender inequality in textbooks worldwide.

# Citations